\pdfoutput=1
\documentclass[11pt]{article}

\usepackage[preprint]{acl}

\usepackage{times}
\usepackage{latexsym}

\usepackage[T1]{fontenc}
\usepackage[utf8]{inputenc}

\usepackage{microtype}

\usepackage{inconsolata}

\usepackage{graphicx}

\usepackage{amsthm}
\usepackage[most]{tcolorbox}
\usepackage{fancyvrb}
\usepackage{fvextra}
\usepackage{booktabs,multirow}

\newcommand{\ie}{\textit{i.e.}}
\newcommand{\eg}{\textit{e.g.}}

\newcommand{\cD}{\mathcal{D}}

\newcommand{\cL}{\mathcal{L}}

\newcommand{\cR}{\mathcal{R}}

\newcommand{\tI}{\tilde{I}}
\newcommand{\tQ}{\tilde{Q}}
\newcommand{\tR}{\tilde{R}}

\newtheorem{example}{Example}

\title{Logic-of-Thought: Empowering Large Language Models with \\Logic Programs for Solving Puzzles in Natural Language}

\author{Naiqi Li$^1$,
Peiyuan Liu$^1$,
Zheng Liu$^1$,
Tao Dai$^{2}$,
Yong Jiang$^{1}$,
Shu-Tao Xia$^{1}$\\
$^1$Tsinghua Shenzhen International Graduate School\quad
$^2$Shenzhen University\\
\{linaiqi.thu, daitao.edu\}@gmail.com, 
\{lpy23, liu-z24\}@mails.tsinghua.edu.cn\\
\{jiangy, xiast\}@sz.tsinghua.edu.cn
 \\}

\begin{document}
\maketitle
\begin{abstract}
Solving puzzles in natural language poses a long-standing challenge in AI. While large language models (LLMs) have recently shown impressive capabilities in a variety of tasks, they continue to struggle with complex puzzles that demand precise reasoning and exhaustive search. In this paper, we propose Logic-of-Thought (Logot), a novel framework that bridges LLMs with logic programming to address this problem. Our method leverages LLMs to translate puzzle rules and states into answer set programs (ASPs), the solution of which are then accurately and efficiently inferred by an ASP interpreter. 
This hybrid approach combines the natural language understanding of LLMs with the precise reasoning capabilities of logic programs. 
We evaluate our method on various grid puzzles and dynamic puzzles involving actions, demonstrating near-perfect accuracy across all tasks.
Our code and data are available at: \url{https://github.com/naiqili/Logic-of-Thought}.
\end{abstract}

\section{Introduction}

Solving puzzles expressed in natural language has long been regarded as a cornerstone task for artificial intelligence, which requires a diverse set of advanced skills, including natural language understanding, logical reasoning, abstract thinking, and the ability to plan and infer~\cite{mittal2024puzzlebench,li2024assessing}.

In recent years, large language models (LLMs) such as GPT~\cite{brown2020language}, PaLM~\cite{chowdhery2023palm}, and LLaMA~\cite{touvron2023llama} have shown remarkable progress in tasks like numerical and commonsense reasoning. These models have demonstrated impressive capabilities in few-shot and zero-shot learning, which lead to the development of several prompting strategies including Chain-of-Thought (CoT)~\cite{wei2022chain}, Tree-of-Thought (ToT)~\cite{yao2023tree}, and Program-of-Thought (PoT)~\cite{chen2023program}. 
These prompting techniques aim to guide LLMs to reason more systematically, and have further enhanced LLMs' ability in handling simple tasks.

\begin{figure}[!t]
\begin{tcolorbox}[
  title=Puzzle Example: Hitori,
  colback=white,
  colframe=black,
  fonttitle=\bfseries\large,
  coltitle=black,
  sharp corners=south, % for a stylized top-rounded box
  boxrule=0.8pt,
  % drop shadow south east,
  enhanced,
  attach boxed title to top center={yshift=-2mm},
  boxed title style={
    colback=white,
    colframe=black,
    rounded corners
  }
]
% \begin{tcolorbox}[colback=gray!10,
%   colframe=black,
%   boxrule=0.4pt,
%   rounded corners,
%   sharp corners=all,
%   interior hidden,
%   arc=2mm,
%   left=3mm,right=3mm,top=1mm,bottom=1mm
% ]
{
\small
\textbf{Rule Specification:} Hitori is a puzzle on a grid filled with numbers. The objective is to eliminate duplicates by marking some cells as black in each row and column according to following rules: 1) Eliminate numbers by marking them black, so that no row or column has duplicate numbers; 2) Blackened cells cannot be horizontally or vertically adjacent; 3) Not blackened cells must form a single connected group.\\
\textbf{Question Instance}:
\begin{center}
\includegraphics[width=0.7\linewidth]{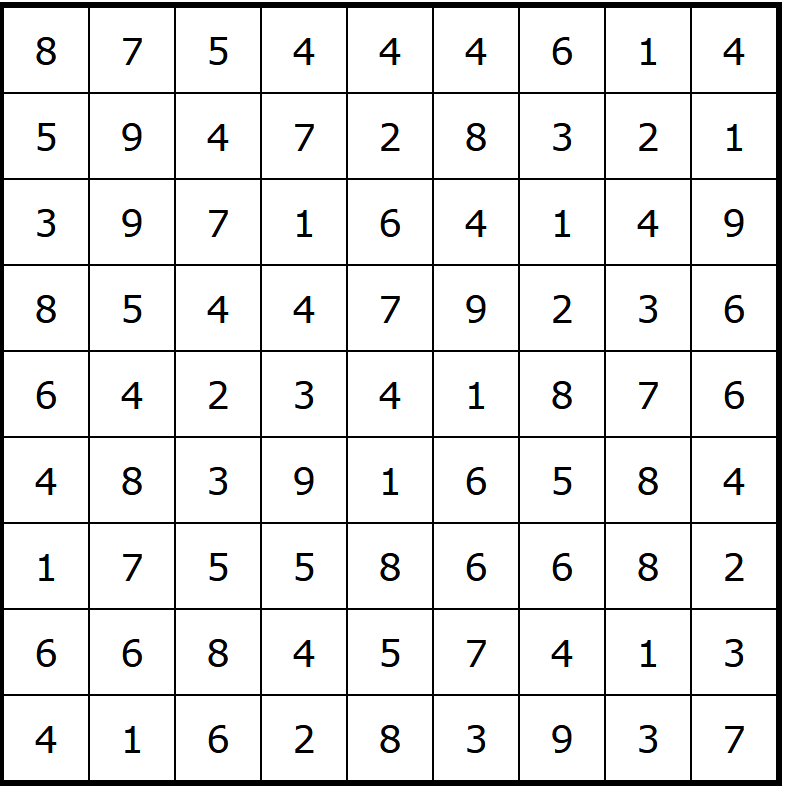}
\end{center}
\textbf{Answer Instance}:
\begin{center}
\includegraphics[width=0.7\linewidth]{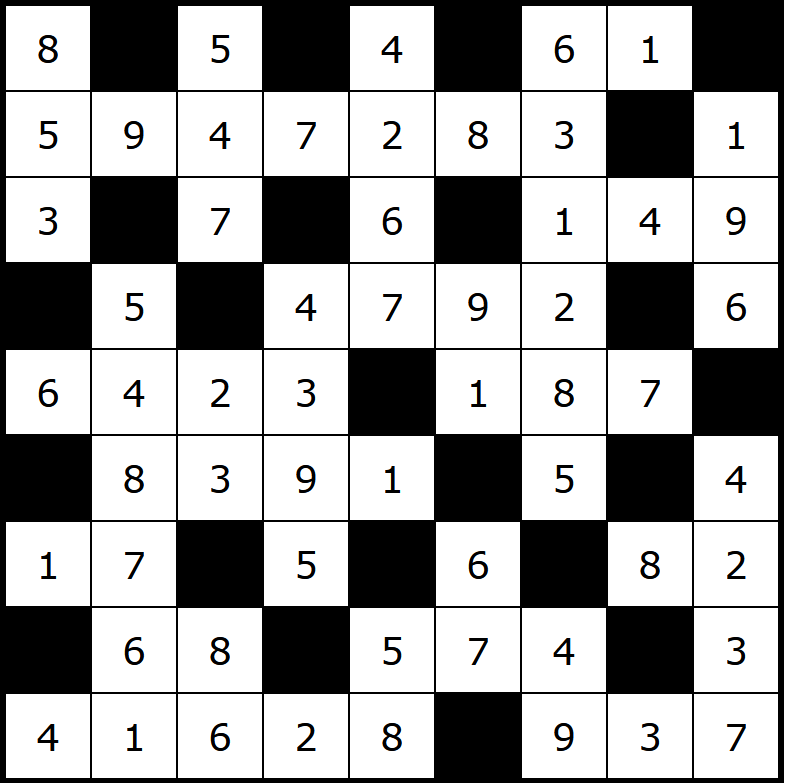}
\end{center}
}
\end{tcolorbox}

    \caption{An example of the Hitori puzzle.}
    \label{fig:intro-hitori}
\end{figure}

However, when facing more involved puzzles such as Hitori in Figure~\ref{fig:intro-hitori}, current LLM-based methods are still inadequate. The puzzles considered in this work require a combination of deep reasoning and precise understanding. Even small misinterpretations can lead to entirely incorrect solutions, as there is little tolerance for ambiguity or error in the problem representation. The need for precise understanding, extensive search, and exact inference exceeds the capabilities of current LLMs and prompting methods. 

To address these challenges, we propose to bridge LLM and logic programming. Logic programs have been well-investigated and are with broad applications in planning, diagnosis, and knowledge representation~\cite{lifschitz2019answer, kowalski2014logic}. A key principle of logic programming is its \textbf{declarative} nature: instead of specifying \textit{how} to solve a problem, one specifies \textit{what} the problem is, which contrasts with popular \textbf{algorithmic} languages.  
The advantages of logic programs include elaboration tolerance (small changes in specification result in minimal changes to the program), separation of knowledge and reasoning algorithms, and the use of powerful off-the-shelf solvers that can perform systematic and exhaustive search~\cite{gelfond2014knowledge}. 
%These characteristics make logic programs particularly suitable for puzzle solving, where precise semantics, rule-based inference, and flexibility in representation are crucial.

In this paper, we propose a novel framework called {Logic-of-Thought} (Logot), which combines the strengths of large language models with the reasoning capabilities of logic programs. Our key idea is to first translate puzzle descriptions into formal logic programs using LLMs. Then logic program interpreters are used to search for solutions with utmost accuracy and high efficiency. This hybrid approach leverages the language understanding and generative capacity of LLMs, while offloading the heavy reasoning and search tasks to well-established logic program engines.

Our main contributions are as follows:
\begin{itemize}
    \item We propose a novel framework to solve puzzles in natural language, which is a task that remains to be a major challenge.
    \item We introduce a new paradigm that integrates LLMs with logic programming. %It leverages elaboration tolerance and separates knowledge representation from reasoning algorithms, enabling accurate and flexible inference.
    \item We curate a diverse set of benchmark puzzles to evaluate our method, including three grid puzzles and four dynamic puzzles involving actions.
    \item Our experiments demonstrate near-perfect accuracy across all puzzle types, showing the effectiveness and generality of our method.
    \item We make our code and dataset publicly available to facilitate future research.
\end{itemize}

\section{Related Work}
\subsection{Puzzle Solving in Computer Science}

The endeavor of employing machines for puzzle solving can date back to Alan Turing, who cracked the Enigma code during World War II. 
In modern research communities, solving puzzles expressed in natural language has long been regarded as a cornerstone task, as it require a diverse set of advanced cognitive skills, including natural language understanding, logical reasoning, abstract thinking, and the ability to plan and infer. Successfully solving such tasks is crucial for showing an agent can approach human-level intelligence~\cite{lake2017building, bisk2020pi}.

Early AI research attempt puzzle solving via formal logic and mathematical modeling. Some recent works still follow this paradigm: \citet{costa2018solving} introduced a tableaux-based deductive system capable of solving Smullyan's puzzles involving propositional logic and first-order logic. \citet{groza2021modelling} illustrated how various logic puzzles can be modeled in first-order logic, and then be solved by mathematical proving tools. \citet{maeda2024mathematical} proposed a mathematical framework to formalize pencil puzzles like Slither and Sudoku, allowing for the application of constraint-based solvers.

Meanwhile, planning tasks that require searching a sequence of actions to transition from an initial state to a goal state, can also be viewed as a type of structured puzzles.
Formal approaches to planning have leveraged logical representations such as STRIPS~\cite{fikes1971strips} and the situation calculus~\cite{reiter2001knowledge}. The Blocks World domain, in particular, has served as a long-standing benchmark for evaluating planning algorithms, due to its simplicity in structure and implications in various reasoning capacities \cite{li2013reasoning,li2015automatic}. 

However, these methods face several practical limitations. 
First, they are tailored to specific puzzle types, making them difficult to generalize. More critically, translating puzzles from natural language into formal representations requires substantial manual effort and expertise.

\subsection{Large Language Models}

Large Language Models (LLMs) have witnessed rapid development in recent years since the introduction of the Transformer architecture~\cite{vaswani2017attention}, which laid the foundation for modern neural language models. 
GPT-2~\cite{radford2019language} marked a significant leap by demonstrating the power of large-scale autoregressive training, which further leads to ChatGPT and GPT-4~\cite{mao2023gpteval,kalyan2024survey}. 
Current LLMs include both commercial models such as Claude and Gemini~\cite{minaee2024large}, as well as open-source models like LLaMA~\cite{touvron2023llama}, Mistral~\cite{jiang2023mistral}, and DeepSeek~\cite{deepseek}. 

A remarkable property of LLMs is their ability to perform in-context learning and few-shot reasoning, \ie, a model can solve unseen tasks based on a few examples provided in the prompt without any parameter updates. 
Chain-of-Thought (CoT)~\cite{wei2022chain} enables models to decompose problems into intermediate reasoning steps. Self-consistency~\cite{wang2022self} aggregates multiple reasoning paths to improve robustness. Least-to-Most prompting~\cite{zhou2022least} guides LLMs through progressive refinement. Tree-of-Thought (ToT)~\cite{yao2023tree} introduces explicit decision paths with branching logic. Most relevant to our work is Program-of-Thought (PoT)~\cite{chen2022program}, which use LLMs to generate algorithmic code (\eg, Python) to solve complex problems. However, our method differs fundamentally: instead of generating procedural programs, we generate declarative logic programs. Declarative programs specify \textit{what} the solution should satisfy, rather than \textit{how} to compute it.

These prompting techniques have been successfully applied to a variety of tasks, including commonsense reasoning and arithmetic reasoning~\cite{cobbe2021training,ling2017program,talmor2019commonsenseqa}. Nevertheless, recent studies have shown that even state-of-the-art LLMs struggle with solving combinatorial and structured puzzles. For instance, \citet{li2024assessing} assessed GPT-4's capability on Minesweeper puzzles and revealed consistent reasoning failures. \citet{mittal2024puzzlebench} proposed PuzzleBench to systematically benchmark LLMs on first-order logic puzzles, concluding that current models still underperform without external tools.

% These findings motivate our work: can we combine the strengths of modern LLMs with formal, symbolic reasoning techniques—particularly logic programming—to create a new paradigm for puzzle solving? By tightly integrating LLMs' natural language understanding with the rigor and modularity of logic programs, our approach aims to bridge this long-standing gap and unlock new potential for high-level cognitive reasoning.

\subsection{Answer Set Programming}

Answer Set Programming (ASP) is a well-established paradigm in the field of declarative programming and non-monotonic reasoning. It has a broad range of applications, due to its expressive power, clear semantics, and availability of efficient solvers~\cite{baral2003knowledge,gelfond2014knowledge,leone2015answer}. %In this work, we adopt ASP as the target logic formalism, where both puzzle rules and question instances are translated into ASP programs, enabling automated reasoning through mature solvers.

An ASP program consists of a set of rules of the form:
\begin{align*}
L_0 \ \texttt{;} & \ ... \ \texttt{;}\ L_k \ \texttt{:-} \\
&\ L_{k+1}, ..., L_m, \texttt{not} \ L_{m+1}, ..., \texttt{not} \ L_n.
\end{align*}
where each $L_i$ is a literal. Intuitively, if all literals in the body are true and none of the negated literals can be proven, then at least one literal in the head must be true. A rule with an empty body is called a \emph{fact}, while a rule with an empty head is a \emph{constraint}. ASP also supports \emph{choice rules}, such as:
\begin{align*}
m \ \{p(X) : q(X)\} \ n \ \texttt{:-} \ L_1, ..., \texttt{not} \ L_n,
\end{align*}
which allows flexible inclusion of subsets of atoms in answer sets.
To solve ASP programs, we employ \textbf{Clingo}~\cite{gebser2011potassco}, a state-of-the-art ASP interpreter that performs both grounding and solving efficiently. % Clingo has been widely used for complex symbolic reasoning tasks and supports advanced ASP features including optimization and aggregates.

ASP has also been applied to puzzle solving~\cite{mitra2015learning}, where handcrafted programs were written to solve logic grid puzzles. However, the types of puzzles addressed are narrow and rely on highly regular logical structures. ASP programs must be manually designed, a process that is often non-trivial and requires deep expertise in both the puzzle and logic programming. %Third, these systems assume structured symbolic inputs and cannot directly handle the natural language descriptions commonly found in puzzle rules and problem instances.

Our work addresses these limitations by proposing a novel method that combines the reasoning power of ASP with the natural language understanding capabilities of large language models. Instead of manually writing ASP programs, we leverage the in-context learning ability of LLMs to automatically translate natural language puzzle descriptions into executable ASP logic programs. This ushers in a new paradigm for solving complex puzzles that require precise understanding, deep reasonin, and exhaustive search.

\section{Method}

\subsection{Overview}

\begin{figure*}
    \centering
    \includegraphics[width=\linewidth]{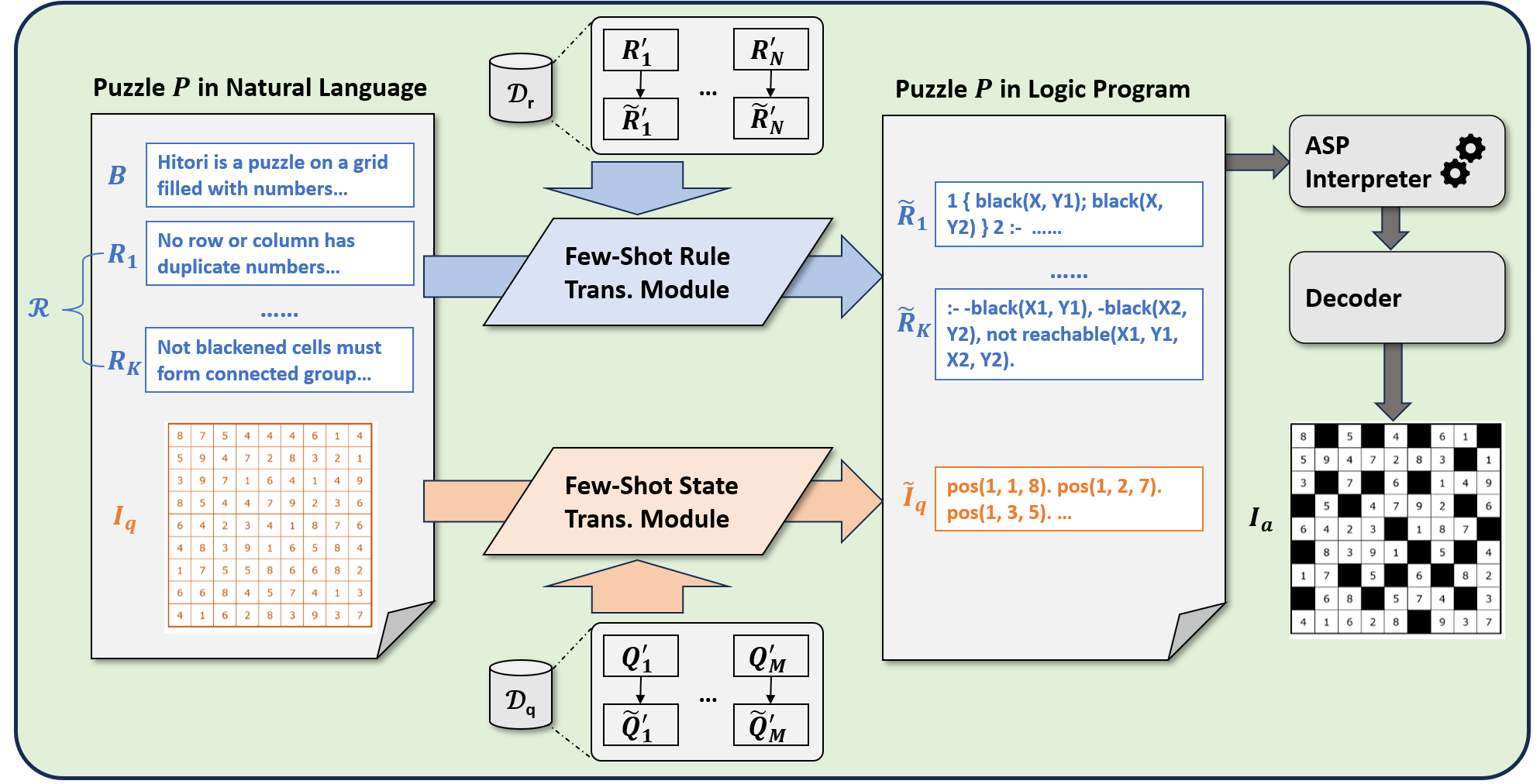}
    \caption{The overall framework of Logot.}
    \label{fig:overview}
\end{figure*}

\textbf{Problem formulation:} We define a puzzle instance as a tuple $P = \langle B, \cR, I_q \rangle$, where $B$ represents background information (\eg, a short textual introduction of the puzzle), $\cR = \{R_i\}_{i=1}^K$ represents a set of rules (\eg, blackened cells cannot be horizontally or vertically adjacent), and $I_q$ represents a particular question instance (\eg, the initial state of the grid). The goal is to design a function $f(P)=I_a$, such that the output answer instance $I_a$ is a proper solution for the puzzle $P$.

Figure~\ref{fig:overview} presents the overall framework of our method.
The key idea is to translate the puzzle's rule specifications and question instance into the corresponding logic program separately. 
Subsequently, off-the-shelves logic programming interpreters are used to ground and search for solutions.
During the translation, the in-context and few-shot learning abilities of large language models are utilized.
To facilitate this, two sets of few-shot examples $\cD_r = \{\langle R_i', \tR_i' \rangle\}_{i=1}^N$ and $\cD_q = \{\langle Q_i', \tQ_i' \rangle\}_{i=1}^M$ are introduced, which correspond to the translations of rule specifications and question instances. Here $\langle R_i', \tR_i' \rangle$ denotes the translation between the rule's textual form and logic program representation (similar for $\langle Q_i', \tQ_i' \rangle$). 

With these notations, the ultimate objective is to design a function $f(P; \cD_r, \cD_q)=I_a$, such that the answer $I_a$ resolves the puzzle $P$.

In what follows, we use $\cL(s)$ to denote the output of the LLM when $s$ is given as input, and use $s_1 | s_2 | ... | s_k$ to denote the concatenation of strings.

\subsection{Few-Shot Rule Specification Translation Module}

% For textual rules $R_i \in \cR$, the few-shot rule specification translation module formally converts them to logic programs as follows:

% \begin{align*}
% & RT(R_i; P,\cD_r)=\cL(R_1'|\tR_1'|...|R_N'|\tR_N'|B|R_i),\\
% & RT(\cR; P,\cD_r)=\{RT(R_i; P,\cD_r)|R_i\in \cR\}_{i=1}^K.
% \end{align*}

% The in-context learning ability allows the LLM to produce logic programs of the rules by performing text completion.

% An example of the translation of rule specifications in the Hitori puzzle is presented in Figure~\ref{fig:trans-rule}.

One of the core components of our framework is the \textit{Few-Shot Rule Specification Translation Module}, which is designed to transform natural language rule descriptions $\cR = \{R_i\}_{i=1}^K$ into formal logic programs $\{\tR_i\}_{i=1}^K$ that can be interpreted and executed by logic programming engines. These rules define the constraints and structure of the puzzle, and are expressed in natural language. %Therefore, their translation into a machine-interpretable formalism is crucial for automated reasoning.

To accomplish this, we exploit the in-context learning capabilities of large language models (LLMs), using a set of few-shot examples $\cD_r = \{\langle R_i', \tR_i' \rangle\}_{i=1}^N$ as demonstrations. Each pair in $\cD_r$ consists of a textual rule $R_i'$ and its corresponding logic representation $\tR_i'$. These demonstrations are drawn from puzzles of similar domains and serve as guidance for translating new rules.

Given a new rule $R_i \in \cR$ and the full puzzle instance $P = \langle B, \cR, I_q \rangle$, the rule translation function is defined as:
\begin{align*}
    RT(R_i; P, \cD_r) &= \cL(R_1'|\tR_1'| \dots |R_N'|\tR_N'| B | R_i), \\
    RT(\cR; P, \cD_r) &= \{RT(R_i; P, \cD_r) \mid R_i \in \cR\}_{i=1}^K,
\end{align*}
where $\cL(\cdot)$ denotes the output of the LLM given the concatenated string of examples and the rule $R_i$ to be translated. The inclusion of the background description $B$ provides additional context about the specific puzzle domain, helping the model ground its understanding of the rule semantics.

\begin{figure}[!t]
\centering    
\begin{tcolorbox}[colback=gray!5, colframe=black,fontupper=\small]
------(Few-shot examples from other puzzles)------\\
$R_1'$: Fill each row with the numbers 1 through 9 without repeating any number.\\
$\tR_1'$: :- pos(X, Y1, N), pos(X, Y2, N), Y1 != Y2.\\
$R_2'$: Fill each column with the numbers 1 through 9 without repeating any number.\\
$\tR_2'$: :- pos(X1, Y, N), pos(X2, Y, N), X1 != X2.\\
......\\
-----------------------------------
-----------------------------\\
$B$: Hitori is a puzzle on a grid filled with ...\\
$R_i$: Eliminate numbers by marking them black, so no row or column has duplicate numbers.\\
--------------------(LLM output, $\tR_i$)--------------------\\
:- pos(X, Y1, N), pos(X, Y2, N), Y1 != Y2, not black(X, Y1), not black(X, Y2).\\
:- pos(X1, Y, N), pos(X2, Y, N), X1 != X2, not black(X1, Y), not black(X2, Y).
\end{tcolorbox}
\caption{An example of rule translation in the Hitori puzzle.}
\label{fig:trans-rule}
\end{figure}

This translation process allows the LLM to infer the appropriate formal structure for a rule based on analogical reasoning from the provided demonstrations. An example of such a translation is shown in Figure~\ref{fig:trans-rule}.

By leveraging LLMs for rule translation, Logot presents a novel method for converting human-readable rules into executable logic, serving as a critical bridge between natural language and symbolic reasoning in our system.

\subsection{Few-Shot Puzzle State Translation Module}

\begin{figure}
\centering    
\begin{tcolorbox}[colback=gray!5, colframe=black,fontupper=\small]
$Q_1'$: 112435 115446 ...\\
$\tQ_1'$: 
pos(1, 1, 1). pos(1, 2, 1). pos(1, 3, 2).
pos(1, 4, 4). pos(1, 5, 3). pos(1, 6, 5).
pos(2, 1, 1). pos(2, 2, 1). pos(2, 3, 5). ...\\
......\\
-----------------------------------
-----------------------------\\
$I_q$: 875444614 594728321 ...\\
--------------------(LLM output, $\tI_q$)--------------------\\
pos(1, 1, 8). pos(1, 2, 7). pos(1, 3, 5).
pos(1, 4, 4). pos(1, 5, 4). pos(1, 6, 4).
pos(1, 7, 6). pos(1, 8, 1). pos(1, 9, 4). pos(2, 1, 5). pos(2, 2, 7). pos(2, 3, 4). ...
\end{tcolorbox}
\caption{An example of puzzle state translation in the Hitori puzzle.}
\label{fig:trans-state}
\end{figure}
The \textit{Few-Shot Puzzle State Translation Module} is responsible for converting the puzzle’s initial state $I_q$ into ASP statements $\tI_q$. 

Given a puzzle instance $P = \langle B, \cR, I_q \rangle$ and a set of few-shot examples $\cD_q = \{\langle Q_i', \tQ_i' \rangle\}_{i=1}^M$ from the same puzzle domain, where $Q_i'$ is a textual representation of a puzzle’s initial state and $\tQ_i'$ is its corresponding logic encoding, the module generates the translated logic program $\tI_q$ by prompting the LLM as follows:
\begin{align*}
    ST(I_q; P, \cD_q) = \cL(Q_1'|\tQ_1'| \dots |Q_M'|\tQ_M'| I_q).
\end{align*}
Here $\cL$ denotes the output of the language model given the prompt, which is constructed from concatenating $M$ few-shot examples and the input $I_q$.

This formulation leverages the LLM's ability to learn translation patterns from a small number of demonstration pairs.
An illustrative example of this translation process in the Hitori puzzle is shown in Figure~\ref{fig:trans-state}. 

\subsection{Inference and Postprocessing}

Once the puzzle rules and the question instance have been translated into their respective logic program representations, the next step is to perform automated reasoning to obtain a solution. 

Specifically, we combine the outputs of the \textit{Rule Specification Translation Module} and the \textit{Puzzle State Translation Module} to form a complete answer set programming specification. 
We then invoke a state-of-the-art ASP solver, Clingo~\cite{gebser2011potassco}, to compute the answer sets that satisfy all constraints in the program. 
Finally, a lightweight postprocessing component $D(\cdot)$ is applied to convert the ASP output into a human-readable answer instance $I_a$. This decoder interprets the logic atoms in the answer set (e.g., \texttt{black(2,3)}, \texttt{pos(1,1,5)}) and formats them into a structured representation consistent with the puzzle’s original format.

Formally, the overall pipeline of our method can be represented as:
\begin{align*}
&I_a = f(P;\cD_r, \cD_q) = \\
&D\left(ASP\left(RT\left(\cR; P,\cD_r\right) \cup \{ST(I_q; P,\cD_q)\}\right)\right),
\end{align*}
where $RT(\cR; P, \cD_r)$ denotes the translated set of puzzle rules, and $ST(I_q; P, \cD_q)$ represents the logic encoding of the initial puzzle state. The union of these two components yields a complete ASP program that fully specifies the puzzle constraints and configuration.

To summarize, the key advantage of our method lies in its ability to seamlessly integrate the natural language understanding capabilities of large language models with the rigorous reasoning framework of logic programming, establishing a new paradigm for puzzle solving and reasoning.

\begin{table*}[t]
\centering
\small
\begin{tabular}{l|l|ccc|cccc}
\toprule
\multirow{2}{*}{Method} & \multirow{2}{*}{Model} & \multicolumn{3}{c|}{{Classic Grid Puzzles}} & \multicolumn{4}{c}{{Dynamic Puzzles with Actions}} \\
% \cmidrule{c}{3-9}
& & Sudoku & Hitori & Fillomino & BW-GR & BW-LG & BW-PV & BW-PJ \\
\midrule
{Standard } & Deepseek-V3     & 0\%   & 0\%   & 0\%   & 59.0\% & 88.5\% & 79.0\% & 88.5\% \\
                                Prompt & GPT-4o-mini     & 0\%   & 0\%   & 0\%   & 47.5\% & 50.5\% & 71.0\% & 85.0\% \\
\midrule
\multirow{1}{*}{CoT}             & Deepseek-V3     & N/A   & N/A   & N/A   & 62.5\% & 97.0\% & 97.0\% & 100.0\% \\
\midrule
\multirow{2}{*}{Finetune*}       & RoBERTa         & N/A   & N/A   & N/A   & 96.8\% & 99.7\% & 87.6\% & 87.4\% \\
                                 & GPT-2           & N/A   & N/A   & N/A   & 97.4\% & 99.4\% & 90.1\% & 85.1\% \\
\midrule
\multirow{3}{*}{Logot}           & Deepseek-V3     & 99.0\% & \textbf{100.0\%} & 91.0\% & 92.0\% & 92.5\% & 98.0\% & 98.0\% \\
                                 & GPT-4o-mini     & \textbf{100.0\%} & \textbf{100.0\%} & \textbf{100.0\%} & 95.5\% & 99.5\% & 98.5\% & 95.0\% \\
                                 & GPT-4o          & \textbf{100.0\%} & \textbf{100.0\%} & \textbf{100.0\%} & \textbf{97.5\%} & \textbf{100.0\%} & \textbf{100.0\%} & \textbf{100.0\%} \\
\bottomrule
\end{tabular}
\caption{Accuracy of different methods on grid puzzles (Sudoku, Hitori, Fillomino) and dynamic puzzles in the Blocks World domain (Goal Recognition - GR, Legality - LG, Plan Verification - PV, Projection - PJ).}
\label{tab:accuracy}
\end{table*}

\section{Experiment}

\subsection{Task Description}

We evaluate our method on two types of puzzles: \textit{grid puzzles} and \textit{dynamic puzzles involving actions}.

\paragraph{Grid Puzzles.}
These include \textbf{Sudoku}, \textbf{Hitori}, and \textbf{Fillomino}, which are widely known and require nontrivial combinatorial search. Solving them demands precise rule interpretation and reasoning over a large search space. For each of the three puzzles, we collect 200 instances with corresponding ground-truth solutions.

\paragraph{Dynamic Puzzles.}
We use four tasks from the \textbf{Blocks World} domain introduced in \citet{he2023exploring}: \textit{Projection}, \textit{Legality}, \textit{Plan Verification}, and \textit{Goal Recognition}. The Blocks World consists of an unbounded table surface and a set of blocks that can be stacked to form towers. Each block can either be on another block or directly on the table. Movement of a block is constrained by preconditions, such as whether it is clear.
As an example, the \textit{Legality} task asks whether a given sequence of actions can be executed from an initial state. In the original dataset of \citep{he2023exploring}, each task contains 12,000 training and 1,000 test instances. In our experiment, we randomly sample 200 instances per task for evaluation.

Due to space constraints, detailed descriptions and examples for all puzzles are left in the Appendix.

\subsection{Comparison}

To demonstrate the effectiveness of our method, we compare it against several baselines across a variety of puzzles. 
% The results, summarized in Table~\ref{tab:accuracy}, show that our method consistently outperforms others.

\paragraph{Baselines.} We consider the following baseline methods:

\begin{itemize}
    \item \textbf{Standard Prompting:} Directly prompts the language model with the puzzle description and expects the answer.
    \item \textbf{Chain-of-Thought (CoT)}~\cite{wei2022chain}: Prompts the model to generate intermediate reasoning steps before arriving at an answer. 
    \item \textbf{Finetuning:} Finetunes language models on task-specific datasets. We adopt the results reported in \cite{he2023exploring}.
\end{itemize}

Each baseline is evaluated using different LLM backbones, including \texttt{Deepseek-V3}, \texttt{RoBERTa}, \texttt{GPT-2}, \texttt{GPT-4o-mini}, and \texttt{GPT-4o}.

\paragraph{Accuracy Analysis.} The comparison results are shown in Table~\ref{tab:accuracy}. ``N/A'' indicates that the method is not applicable or is difficult to adapt. For example, it is challenging to design effective chain-of-thought (CoT) prompts for grid puzzles such as Sudoku.
The results show that:

\begin{itemize}
    \item {Logot} consistently achieves superior performance across both classic grid puzzles and dynamic puzzles with actions. 
    \item Notably, \texttt{Logot+GPT-4o} achieves near-perfect accuracy on all tasks. 
    \item Even when using smaller or less capable models like \texttt{Deepseek-V3} or \texttt{GPT-4o-mini}, Logot still demonstrates robust performance.
    \item Although CoT performs competitively on some dynamic puzzles (e.g., BW-LG, BW-PJ), such applications were not explored in previous work~\cite{he2023exploring}, suggesting that further investigation into prompting strategies may yield useful insights. An example of the CoT prompts we used is shown in Figure~\ref{fig:cot-bw}.
\end{itemize}

\begin{figure*}
\centering    
\begin{tcolorbox}[colback=gray!5, colframe=black,fontupper=\small]
\begin{Verbatim}[breaklines=true]
1. Identify which block is being moved.
2. Find where that block currently is.
3. Check whether the block is clear (nothing on top of it).
4. Check whether the destination block is clear.
5. If the block is clear, is currently in the stated location, and the destination is clear, then the action is legal. Otherwise, it is not.
6. Conclude with "True" if legal, or "False" if not.

### Example 1:
state:
The red block is on the table. The tan block is on the table. The green block is on the red block. The green block is clear. The tan block is clear.

query:
Jane moves the green block from the red block to the tan block.

Reasoning:
The green block is the one being moved.
It is currently on the red block.
It is clear, meaning there is nothing on top of it.
The tan block is clear as well.
All conditions are satisfied to move the green block from the red block to the tan block.

Final output: True

### ......(other examples)
\end{Verbatim}
\end{tcolorbox}
\caption{A chain-of-thought (CoT) prompt example for solving the Blocks World - Legality puzzle.}
\label{fig:cot-bw}
\end{figure*}

\begin{table*}[t]
\centering
\small
\begin{tabular}{l|l|ccc|cccc}
\toprule
\multirow{2}{*}{Method} & \multirow{2}{*}{Model} & \multicolumn{3}{c|}{{Classic Grid Puzzles}} & \multicolumn{4}{c}{{Dynamic Puzzles with Actions}} \\
% \cmidrule{c}{3-9}
& & Sudoku & Hitori & Fillomino & BW-GR & BW-LG & BW-PV & BW-PJ \\
\midrule
{Standard } & Deepseek-V3     & 0.0048 & 0.0089 & 0.0087 & 0.0119 & 0.0101 & 0.0120 & 0.0105 \\
                                Prompt & GPT-4o-mini     & 0.0073 & 0.0113 & 0.0106 & 0.0129 & 0.0109 & 0.0130 & 0.0114 \\
\midrule
\multirow{1}{*}{CoT}             & Deepseek-V3     & N/A   & N/A   & N/A   & 0.0895 & 0.0303 & 0.0957 & 0.0629 \\
\midrule
\multirow{3}{*}{Logot}           & Deepseek-V3    & 0.0920 & 0.0514 & 0.0393 & 0.0474 & 0.0418 & 0.0478 & 0.0506 \\
                                 & GPT-4o-mini     & 0.1619 & 0.0800 & 0.0507 & 0.0607 & 0.0542 & 0.0620 & 0.0654 \\
                                 & GPT-4o          & 4.9572 & 2.4568 & 1.5565 & 1.8659 & 1.6618 & 1.9028 & 2.0023
 \\
\bottomrule
\end{tabular}
\caption{Cost of solving each puzzle type across different methods and models (in USD). }
\label{tab:cost}
\end{table*}

\begin{figure}
    \centering
    \includegraphics[width=0.9\linewidth]{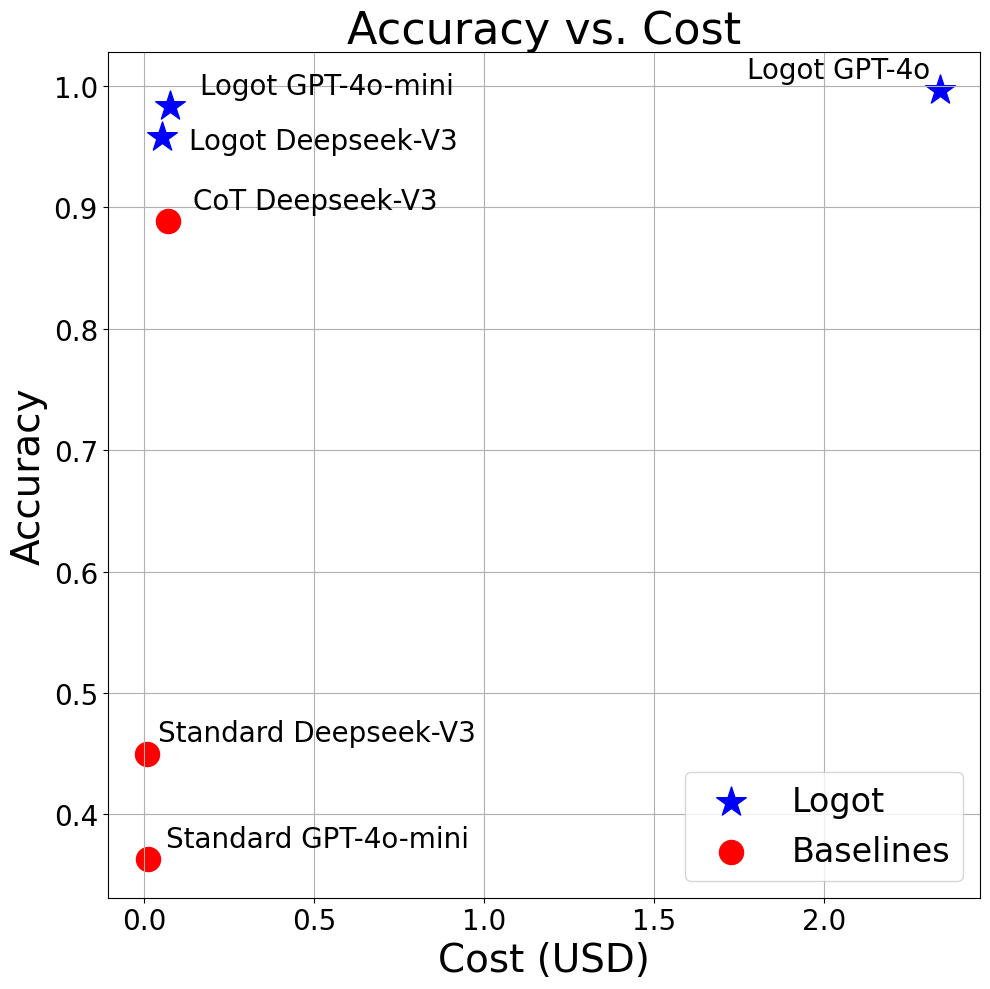}
    \caption{Accuracy-cost tradeoff of different methods.}
    \label{fig:acc_cost}
\end{figure}

\paragraph{Cost Analysis.} 
We further analyze the cost-efficiency tradeoff of different methods, based on Table~\ref{tab:cost} and Figure~\ref{fig:acc_cost}.

Table~\ref{tab:cost} reports the total cost (in USD) required to solve 200 instances for each puzzle category. %For each method, the cost is computed by aggregating the expenses over all test puzzles across the seven datasets (three classic grid puzzles and four dynamic puzzles). 
Figure~\ref{fig:acc_cost} provides a clearer summary by illustrating the trade-off between cost and accuracy averaged over all instances in each puzzle category.

From the results, we observe that our proposed method is highly cost-effective. When paired with GPT-4o, {Logot} achieves near-perfect accuracy (over 99\%) across all puzzles. Although this comes at a relatively higher cost, the performance gain is significant.

More importantly, {Logot} can also be deployed with more affordable language models, such as Deepseek-V3 or GPT-4o-mini. In these settings, it still achieves strong accuracy (over 95\% and 98\%, respectively), while incurring significantly lower costs. 
% This demonstrates the versatility and efficiency of our method.
% These results highlight the key advantage of our approach: by effectively leveraging the structure and reasoning capabilities of logic programs, \textsc{Logot} can drastically reduce the reliance on expensive and powerful LLMs. 
All evidences suggest that, Logot is not only highly accurate but also resource-efficient, making it practical and accessible for solving various challenging puzzles.

\subsection{Failure Case Study}

\begin{figure}
\centering    
\begin{tcolorbox}[colback=gray!5, colframe=black,fontupper=\small]
Rule Specification: Sudoku is a puzzle played on a grid...

Init State: Now solve the puzzle instance presented below:

\begin{center}
050023064\\
104000000\\
860000000\\
905000002\\
000507000\\
400000900\\
090400800\\
006700010\\
0000000\textcolor{red}{59}
\end{center}

Translated ASP Program:\\
......\\
pos(1, 2, 5). pos(1, 5, 2). pos(1, 6, 3).\\ pos(1, 8, 6). pos(1, 9, 4). pos(2, 1, 1).\\ pos(2, 3, 4). pos(3, 1, 8). pos(3, 2, 6).\\ pos(4, 1, 9). pos(4, 3, 5). pos(4, 9, 2).\\ pos(5, 4, 5). pos(5, 6, 7). pos(6, 1, 4).\\ pos(6, 7, 9). pos(7, 2, 9). pos(7, 4, 4).\\ pos(7, 7, 8). pos(8, 3, 6). pos(8, 4, 7).\\ pos(8, 8, 1). \textcolor{red}{pos(9, 7, 5). pos(9, 8, 9)}.
\end{tcolorbox}
\caption{A failure case of the Sudoku puzzle.}
\label{fig:failure-sudoku}
\end{figure}

\begin{figure}
\centering    
\begin{tcolorbox}[colback=gray!5, colframe=black,fontupper=\small]
Rule Specification: In the Blocks World domain...

Init State: The silver block is on top of the teal block. The green block is on top of the indigo block. The silver block is clear. The indigo block is on the table. The teal block is on the table. The green block is clear. The aquamarine block is clear. The aquamarine block is on the table.

Action Sequence: \textcolor{red}{Jane moves the aquamarine block from the teal block to the silver block.}

Translated ASP Program:\\
......\\
\% State Encoding:\\
holds(on(silver, teal), 0).\\
holds(on(green, indigo), 0).\\
holds(clear(silver), 0).\\
holds(on(indigo, table), 0).\\
holds(on(teal, table), 0).\\
holds(clear(green), 0).\\
holds(clear(aquamarine), 0).\\
holds(on(aquamarine, table), 0).\\
\% All Blocks:\\
block(silver). block(green). block(indigo). block(teal). block(aquamarine).\\
\% Query Encoding:\\
\textcolor{red}{occurs(move(aquamarine, table, silver), 0).}\\
\% Steps:\\
\#const num\_step=1.
\end{tcolorbox}
\caption{A failure case of the Blocks World - Legality puzzle.}
\label{fig:failure-bw-lg}
\end{figure}

In this section, we analyze representative failure cases of our method to better understand its limitations and potential areas for improvement. Specifically, we focus on two puzzle categories: Sudoku (Figure~\ref{fig:failure-sudoku}) and Blocks World -- Legality (Figure~\ref{fig:failure-bw-lg}), which are representative of grid puzzles and dynamic puzzles involving actions.

Across all examined cases, we find that the few errorsstem entirely from the \textit{state translation} stage, \ie, incorrect interpretation of the initial state. These mistakes are generally straightforward to identify and are easy to be corrected by humans.

Interestingly, even with powerful LLMs such as \texttt{GPT-4o}, these seemingly simple translation errors can still occur. This highlights a key challenge in achieving full robustness: while logic interpreters are precise and reliable, they depend heavily on accurate input representations, making the translation step a critical bottleneck.

To address this, one promising direction is to integrate more advanced prompting methods such as \textit{Program-of-Thought} (PoT), which uses language models to generate intermediate algorithmic code (\eg, Python) to assist with the transaction of rules and states. By combining PoT with our pipeline, it is possible to further enhance translation accuracy and eliminate residual errors, which we leave for future investigation.

\section{Conclusion}
In this paper, we proposed {Logic-of-Thought} (Logot), a novel framework that combines large language models with logic programming to solve puzzles expressed in natural language, which remains a challenging task. Our method leverages the few-shot in-context learning capabilities of LLMs to translate both puzzle rules and instances into declarative logic programs, which are then executed using an Answer Set Programming interpreter to accurately and efficiently infer solutions.
Experiments show that Logot achieves near-perfect accuracy across a range of puzzle at a reasonable cost. Overall, Logot introduces a new paradigm that combines the strengths of LLM and logic programming, with potential applications extending well beyond the puzzle domain.

\section*{Limitations}

While our proposed method demonstrates strong performance across a range of puzzle types, we acknowledge a few minor limitations that open up promising directions for future research.
First, the number of puzzle types and categories evaluated in this work is relatively modest. 
Future work can further enhance generality by curating a broader and richer collection of puzzles, particularly by exploring more tasks in the planning domain.
Second, our method currently relies on manually annotated few-shot examples to bootstrap performance, which is a common setup in in-context learning. 
Nonetheless, future research could explore the information retrieval techniques~\citet{rubin2022learning} to automatically retrieve high-quality examples, potentially reducing the annotation effort even further.
Finally, occasional errors still occur during the puzzle state translation stage. The accuracy of our method can be further improved using advanced prompting techniques such as Program-of-Thought (PoT), or by leveraging diagnostic tools from the logic programming community.

% \section*{Acknowledgments}

% Bibliography entries for the entire Anthology, followed by custom entries
%\bibliography{anthology,custom}
% Custom bibliography entries only
\bibliography{main}

\newpage
\appendix

\section{Task Description}

\subsection{Classic Grid Puzzles}

\subsubsection*{Sudoku}

Sudoku is a popular logic-based number placement puzzle. The puzzle is played on a 9×9 grid divided into nine 3×3 subgrids (also called regions, boxes, or blocks). The goal is to fill in the grid so that each number appears exactly once in each row, column, and 3×3 subgrid. Specifically, the puzzle requires completing grid under the following rules:
\begin{enumerate}
    \item Fill each row with the numbers 1 through 9 without repeating any number.
    \item Fill each column with the numbers 1 through 9 without repeating any number.
    \item Fill each 3×3 subgrid with the numbers 1 through 9 without repeating any number.
\end{enumerate}

\begin{figure*}[t]
\includegraphics[width=0.48\linewidth]{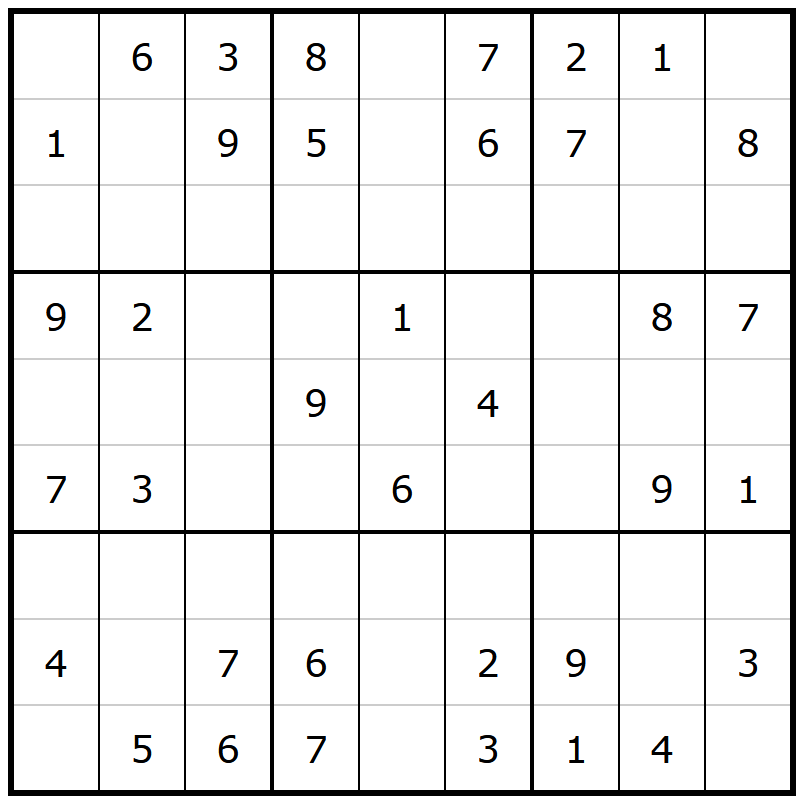} \hfill \includegraphics[width=0.48\linewidth]{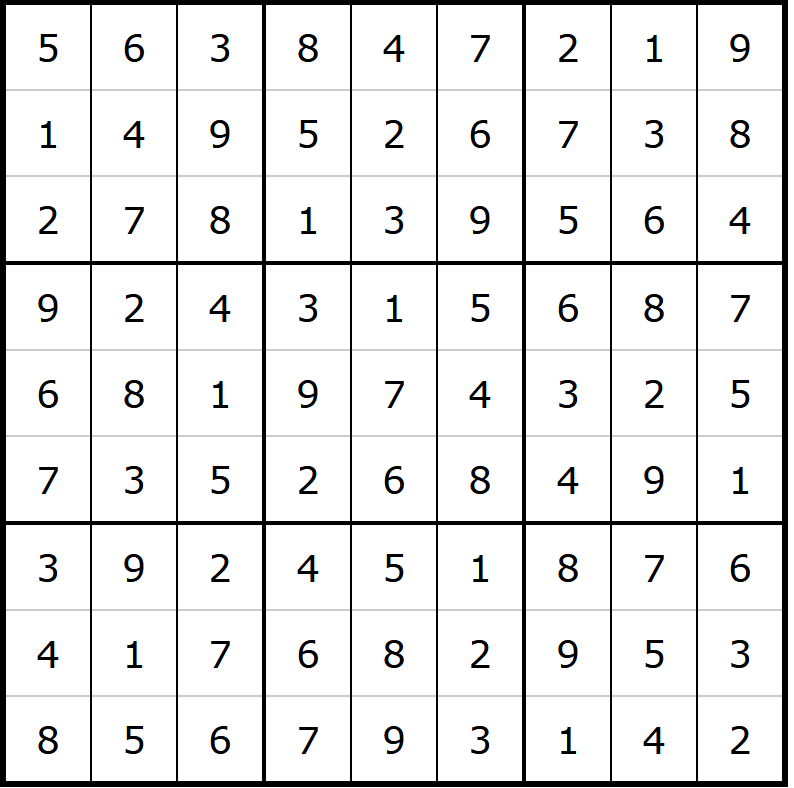}
\caption {An example of the Sudoku puzzle. (Left) Puzzle question; (Right) Puzzle answer.}
\label{fig:example_sudoku}
\end{figure*}

Figure \ref{fig:example_sudoku} presents a question-answer pair of the Sudoku puzzle.

\subsubsection*{Hitori}

Hitori is a logic-based puzzle game originating from Japan. The name ``Hitori'' means ``alone'' or ``one person'' in Japanese, reflecting the puzzle’s goal of isolating numbers. It is typically played on a square grid filled with numbers, and the objective is to eliminate duplicates in each row and column according to following rules:
\begin{enumerate}
    \item Eliminate numbers by marking them (usually shaded or blacked out) so that no row or column has duplicate numbers.
    \item You cannot shade two adjacent cells (cells sharing an edge) — shaded cells must not touch horizontally or vertically.
    \item All unshaded (white) cells must form a single connected group, meaning you can move from any unshaded cell to any other through neighboring unshaded cells.
    \item A shaded cell is considered "eliminated" and cannot be part of the connected group.
\end{enumerate}

\begin{figure*}[t]
\includegraphics[width=0.48\linewidth]{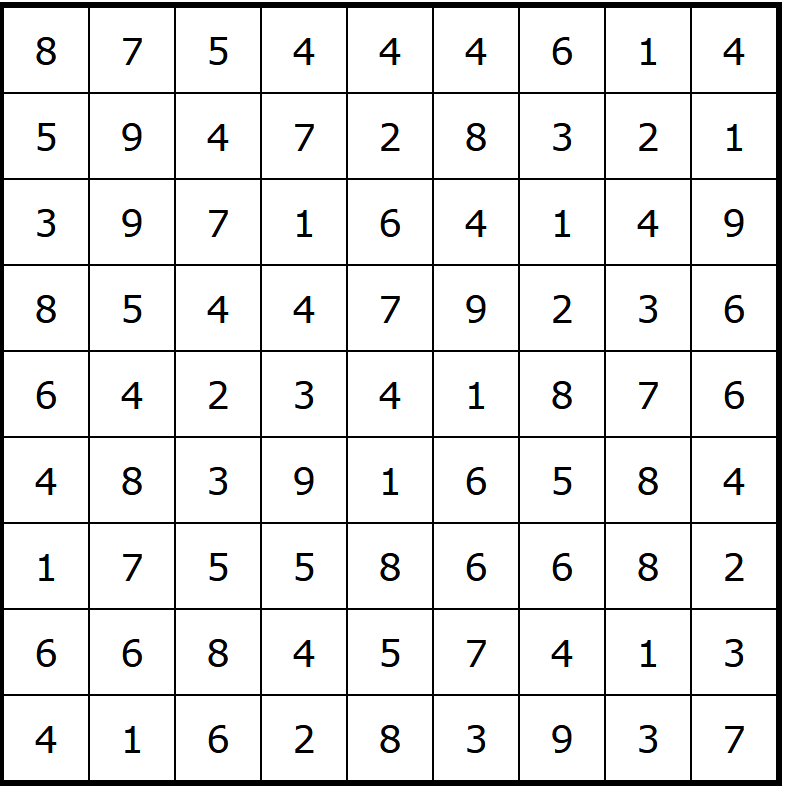} \hfill \includegraphics[width=0.48\linewidth]{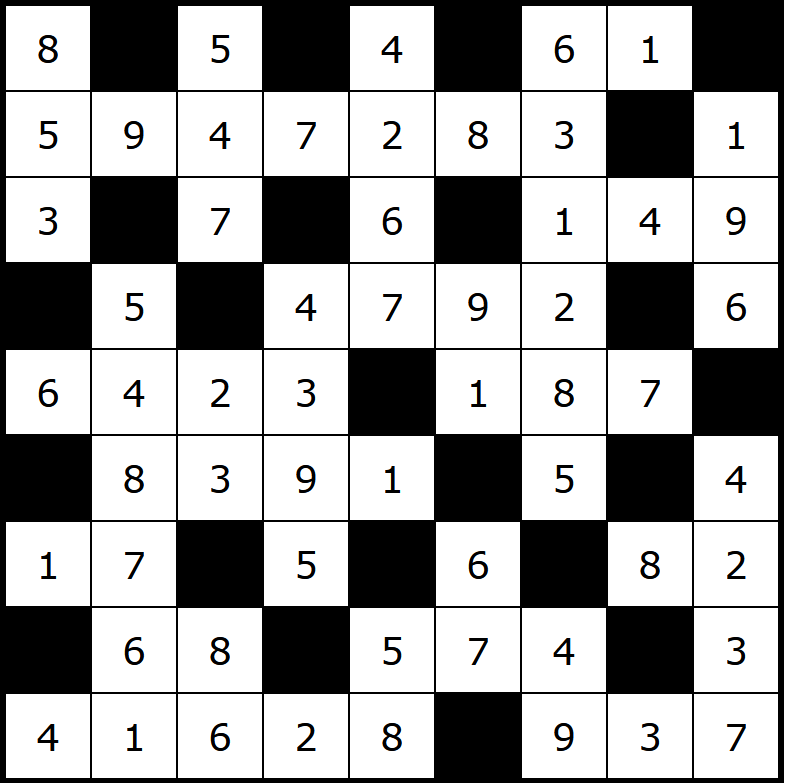}
\caption {An example of the Hitori puzzle. (Left) Puzzle question; (Right) Puzzle answer.}
\label{fig:example_hitori}
\end{figure*}

Figure \ref{fig:example_hitori} presents a question-answer pair of the Hitori puzzle.

\subsubsection*{Fillomino}

Fillomino is a logic puzzle played on a rectangular grid where some cells may initially contain numbers. The goal is to divide the grid into regions, or "polyominoes," such that each region contains exactly one number and has an area (number of cells) equal to that number.
The specific rules are:
\begin{enumerate}
    \item Divide the grid into regions where each region consists of connected cells (horizontally or vertically adjacent).
    \item Each region must contain exactly one number that matches the total number of cells in that region.
    \item Regions of the same size must not be orthogonally adjacent (they cannot share a side).
    \item Empty cells must be filled with numbers during solving to satisfy the above conditions.
\end{enumerate}

\begin{figure*}[t]
\includegraphics[width=0.48\linewidth]{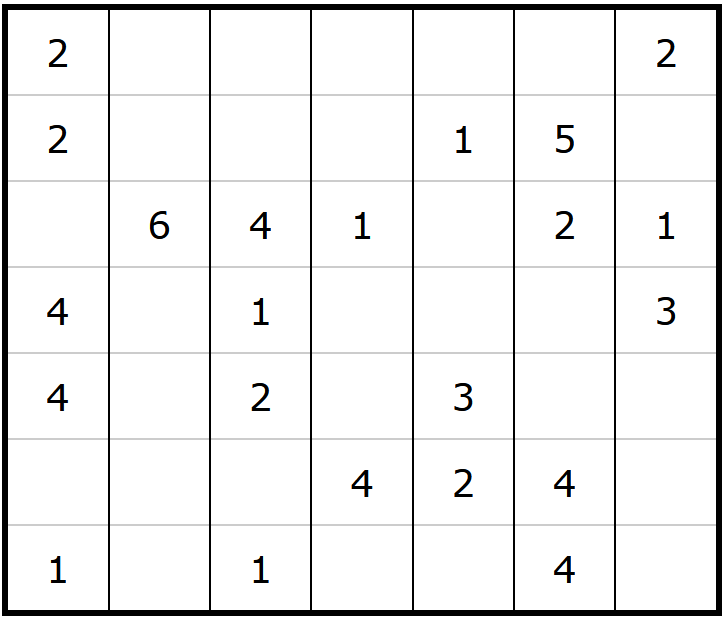} \hfill \includegraphics[width=0.48\linewidth]{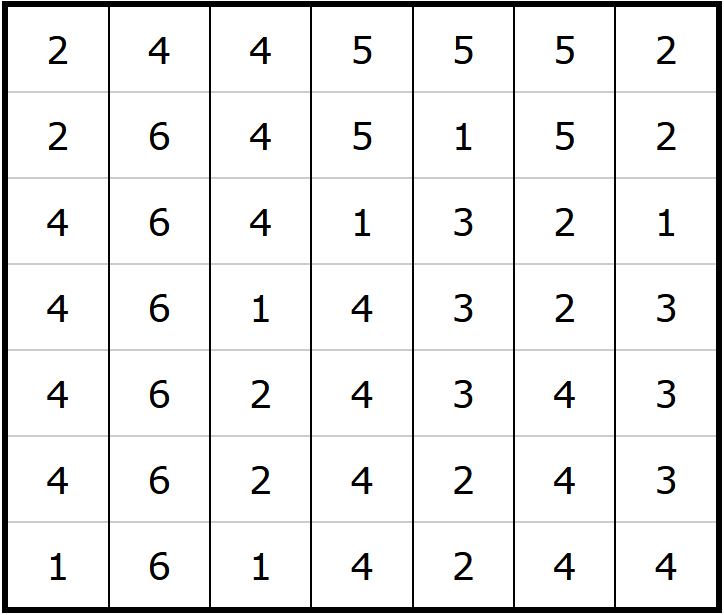}
\caption {An example of the Fillomino puzzle. (Left) Puzzle question; (Right) Puzzle answer.}
\label{fig:example_fillomino}
\end{figure*}

Figure \ref{fig:example_fillomino} presents a question-answer pair of the Fillomino puzzle.

\subsection{Task with Actions}

\textit{Blocks World} is a classic domain in knowledge representation and action reasoning. It involves a set of blocks placed on a table, where blocks can be stacked or moved by an agent under certain physical constraints.

Specifically, the domain is formalized using the following fluents—state properties that may change over time due to actions—and a set of corresponding actions:

\textbf{Fluents}:
\begin{itemize}
    \item \texttt{on(x, y)}: block \texttt{x} is on top of \texttt{y}, where \texttt{y} is a location that can be a block or the table.
    \item \texttt{clear(x)}: block \texttt{x} has nothing on top of it.
\end{itemize}

\textbf{Actions}:
\begin{itemize}
    \item \texttt{move(x, y, z)}: move a block \texttt{x} from location \texttt{y} to location \texttt{z}.
\end{itemize}

Our dataset is based on the recent work of \citet{he2023exploring}, which introduces a structured and large-scale Blocks World benchmark for exploring the capacity of pretrained language models for reasoning about actions and change. 
We consider the four sub-tasks proposed in their paper, \ie, projection, legality, plan verification, and goal recognition.

\subsubsection*{Projection}

The projection task focuses on reasoning about the outcomes of actions. It involves determining whether a given proposition will be true after performing a specific sequence of actions starting from an initial state.

\begin{example}[Projection Task]
The following is an example of negative cases, \ie, the query proposition does not hold after performing the action sequence from the given initial state.
\begin{itemize}
    \item State: The tan block is on the table. The turquoise block is clear. The teal block is on the table. The brown block is clear. The green block is clear. The brown block is on the table. The green block is on the table. The tan block is clear. The turquoise block is on top of the teal block.
    \item Action sequence: Jane moves the turquoise block from the teal block to the brown block. Jane moves the turquoise block from the brown block onto the table. Jane moves the tan block from the table to the turquoise block.
    \item Query: The teal block is on top of the brown block. The green block is clear.
    \item Label: False.
\end{itemize}
\end{example}

\subsubsection*{Legality}

This task focuses on action preconditions. It requires determining whether a given sequence of actions can be executed in order, starting from an initial state, without violating any preconditions.

\begin{example}[Legality Task]
The following is an example of negative cases, \ie, the query action sequence is not executable in the given state.
\begin{itemize}
    \item State: The tan block is on the table. The purple block is clear. The purple block is on the table. The pink block is clear. The blue block is clear. The green block is clear. The blue block is on the table. The green block is on the table. The tan block is clear. The pink block is on the table.
    \item Query: Jane moves the green block from the blue block onto the table. Jane moves the green block from the purple block to the blue block. Jane moves the pink block from the green block to the blue block.
    \item Label: False.
\end{itemize}
\end{example}

\subsubsection*{Plan Verification}

Planning involves identifying a sequence of actions that leads to a desired outcome. In this task, the focus is on verifying whether a given sequence of actions, when applied to an initial state, successfully results in the intended goal.

\begin{example}[Plan Verification Task]
The following is an example of negative cases, \ie, starting from the given state, the query action sequence leads to a state satisfying the goal.
\begin{itemize}
    \item State: The red block is clear. The navy block is on top of the silver block. The red block is on the table. The violet block is on top of the navy block. The silver block is on the table. The magenta block is clear. The violet block is clear. The magenta block is on the table.
    \item Goal: The silver block is not on the table and the red block is on the table.
    \item Query: Jane moves the violet block from the navy block to the magenta block. Jane moves the navy block from the silver block to the red block. Jane moves the silver block from the table to the navy block.
    \item Label: True.
\end{itemize}
\end{example}

\subsubsection*{Goal Recognition}

This task focuses on identifying the intended goal based on a partial observation of actions. Starting from an initial state and given a candidate goal along with an observed sequence of actions, the objective is to determine whether the observed actions are consistent with pursuing that goal—specifically, whether they form the beginning of an optimal plan to achieve it.

\begin{example}[Goal Recognition Task]
The following is an example of negative cases, \ie, the observed sequence of actions is not a prefix of any optimal plan that achieves the goal.
\begin{itemize}
    \item State: The tan block is clear. The olive block is on top of the green block. The brown block is on the table. The magenta block is on top of the brown block. The magenta block is clear. The green block is on the table. The tan block is on top of the olive block.
    \item Observations: Jane moves the magenta block from the brown block to the tan block. Jane moves the magenta block from the tan block onto the table. Jane moves the tan block from the olive block onto the table.
    \item Goal: The brown block is not on the table and the olive block is on top of the magenta block.
    \item Label: False.
\end{itemize}
\end{example}

\section{Prompt Examples}

In Figure~\ref{fig:fs-rule-sudoku}-\ref{fig:cot-bw}, we present various prompt examples for Sudoku and Blocks Would - Legality, which are representative puzzles in classic grid puzzles and dynamic puzzles with action.

\begin{itemize}
\item Figure~\ref{fig:fs-rule-sudoku} shows the few-shot learning prompts for translating rule specifications in Sudoku. 
\item Figure~\ref{fig:fs-state-sudoku} shows the few-shot learning prompts for translating puzzle state in Sudoku. 
\item Figure~\ref{fig:standard-sudoku} shows the standard prompt for solving Sudoku. 
\item Figure~\ref{fig:fs-rule-bw} shows the few-shot learning prompts for translating rule specifications in Blocks World - Legality. 
\item Figure~\ref{fig:fs-state-bw} shows the few-shot learning prompts for translating puzzle state in Blocks World - Legality. 
\item Figure~\ref{fig:standard-bw} shows the standard prompt for solving Blocks World - Legality. 
% \item Figure~\ref{fig:cot-bw} shows the chain-of-thought (CoT) prompt for solving Blocks World - Legality. 
\end{itemize}

\begin{figure*}
\centering    
\begin{tcolorbox}[colback=gray!5, colframe=black,fontupper=\small]
\begin{verbatim}
% Introduction: Hitori is a logic-based puzzle game. 
% It is typically played on a square grid filled with numbers, 
% and the objective is to eliminate duplicates by marking some cells as black
% in each row and column according to specific rules.
% The logic program includes the following atoms:
% coord(X, Y) : The coordination of the grid.
% pos(X, Y, N) : The number in the cell (X, Y) is N.
% num(N) : Possible numbers in the cells, N=1..9
% black(X, Y) : Cell (X, Y) is marked as black.
% adj(X1, Y1, X2, Y2) : Cell (X1, Y1) and cell (X2, Y2) are adjacent to each other.
% reachable(X1, Y1, X2, Y2) : Cell (X1, Y1) and cell (X2, Y2) are reachable to each other.

% Description of a nxn grid.
board_size(n).
coord(1..n,1..n). 
num(1..9).
1 { pos(X, Y, N): num(N)} 1 :- coord(X, Y).

% 1. Eliminate numbers by marking them black,
% so that no row or column has duplicate numbers.
1 { black(X, Y1); black(X, Y2) } 2 :- pos(X, Y1, N), pos(X, Y2, N), Y1 != Y2.
1 { black(X1, Y); black(X2, Y) } 2 :- pos(X1, Y, N), pos(X2, Y, N), X1 != X2.

% Blackened cells cannot be horizontally or vertically adjacent.
:- black(X, Y), black(X + 1, Y).
:- black(X, Y), black(X, Y + 1).

% We have complete knowledge for blackened cells.
-black(X, Y) :- coord(X, Y), not black(X, Y).

% Not blackened cells must form a single connected group, meaning they are reachable 
% to each other, and a cell is not reachable if it is blackened.
adj(X, Y, X, Y+1) :- coord(X, Y), coord(X, Y+1).
adj(X+1, Y, X, Y) :- coord(X, Y), coord(X+1, Y).
adj(X2, Y2, X1, Y1) :- adj(X1, Y1, X2, Y2).
reachable(X1, Y1, X1, Y1) :- -black(X1, Y1).
reachable(X1, Y1, X3, Y3) :- reachable(X1, Y1, X2, Y2), adj(X2, Y2, X3, Y3),
    -black(X1, Y1), -black(X2, Y2), -black(X3, Y3).
reachable(X2, Y2, X1, Y1) :- reachable(X1, Y1, X2, Y2).
:- -black(X1, Y1), -black(X2, Y2), not reachable(X1, Y1, X2, Y2).

......(similar few-shot example for the Fillomino puzzle)

% Introduction: Sudoku is a popular logic-based number placement puzzle. 
% The puzzle is played on a 9×9 grid divided into nine 3×3 subgrids (also called regions).
% The goal is to fill in the grid so that each number appears exactly once in each 
% row, column, and 3×3 subgrid.
% The logic program includes the following atoms:
% coord(X, Y) : The coordination of the grid.
% pos(X, Y, N) : The number in the cell (X, Y) is N.
% num(N) : Possible numbers in the cells, N=1..9
% adj(X1, Y1, X2, Y2) : Cell (X1, Y1) and cell (X2, Y2) are adjacent to each other.
% reachable(X1, Y1, X2, Y2) : Cell (X1, Y1) and cell (X2, Y2) are reachable to each other.

% Description of a 9x9 grid.
ASP_RULES
% Fill each row with the numbers 1 through 9 without repeating any number.
ASP_RULES
% Fill each column with the numbers 1 through 9 without repeating any number.
ASP_RULES
% Fill each 3x3 subgrid with the numbers 1 through 9 without repeating any number.
ASP_RULES
Complete the ASP_RULES.
\end{verbatim}
\end{tcolorbox}
\caption{Prompts for learning rule specifications in the Sudoku puzzles. We first present few-shot examples from other two classic puzzles. Then we prompt the LLM to translate rule specifications from natural language to ASP programs.}
\label{fig:fs-rule-sudoku}
\end{figure*}

\begin{figure*}
\centering    
\begin{tcolorbox}[colback=gray!5, colframe=black,fontupper=\small]
\begin{verbatim}
Input: 
000850000
020000000
010900700
070025093
402000000
000000500
097500000
563000004
000000680
Thought:
pos_(1, 1, 0). pos_(1, 2, 0). pos_(1, 3, 0). pos_(1, 4, 8). pos_(1, 5, 5)...
pos_(2, 1, 0). pos_(2, 2, 2). pos_(2, 3, 0). pos_(2, 4, 0). pos_(2, 5, 0)...
pos_(3, 1, 0). pos_(3, 2, 1). pos_(3, 3, 0). pos_(3, 4, 9). pos_(3, 5, 0)...
pos_(4, 1, 0). pos_(4, 2, 7). pos_(4, 3, 0). pos_(4, 4, 0). pos_(4, 5, 2)...
pos_(5, 1, 4). pos_(5, 2, 0). pos_(5, 3, 2). pos_(5, 4, 0). pos_(5, 5, 0)...
pos_(6, 1, 0). pos_(6, 2, 0). pos_(6, 3, 0). pos_(6, 4, 0). pos_(6, 5, 0)... 
pos_(7, 1, 0). pos_(7, 2, 9). pos_(7, 3, 7). pos_(7, 4, 5). pos_(7, 5, 0)...
pos_(8, 1, 5). pos_(8, 2, 6). pos_(8, 3, 3). pos_(8, 4, 0). pos_(8, 5, 0)...
pos_(9, 1, 0). pos_(9, 2, 0). pos_(9, 3, 0). pos_(9, 4, 0). pos_(9, 5, 0)...
Output:
pos(1, 4, 8). pos(1, 5, 1). 
pos(2, 2, 2). 
pos(3, 2, 1). pos(3, 4, 9). pos(3, 7, 7).
pos(4, 2, 7). pos(4, 5, 2). pos(4, 6, 5). pos(4, 8, 9). pos(4, 9, 3).
pos(5, 1, 4). pos(5, 3, 2).
pos(6, 7, 5). 
pos(7, 2, 9). pos(7, 3, 7). pos(7, 4, 5).
pos(8, 1, 5). pos(8, 2, 6). pos(8, 3, 3). pos(8, 9, 4).
pos(9, 7, 6). pos(9, 8, 8).

Input: 
{input}

In the output, pos(x, y, n) means the number of row x and column y is n. A cell with 0 is 
skipped. Complete the text of Thought and Output. Only show the result, do not explain.
\end{verbatim}
\end{tcolorbox}
\caption{Prompts for translating state representations to ASP programs in the Sudoku puzzle.}
\label{fig:fs-state-sudoku}
\end{figure*}

\begin{figure*}
\centering    
\begin{tcolorbox}[colback=gray!5, colframe=black,fontupper=\small]
\begin{verbatim}
You are a Sudoku-solving assistant. Given a 9x9 Sudoku puzzle, where each row is 
a string of 9 digits and '0' represents an empty cell, solve the puzzle and print 
the completed Sudoku grid.
Input: 
{input}
\end{verbatim}
\end{tcolorbox}
\caption{A standard prompt for solving the Sudoku puzzle.}
\label{fig:standard-sudoku}
\end{figure*}

% \subsection{Examples for Puzzles with Actions}

\begin{figure*}
\centering    
\begin{tcolorbox}[colback=gray!5, colframe=black,fontupper=\small]
\begin{verbatim}% Introduction: Domain: We use a variant of the blocks world (BW) as an example where...
% Task (Goal-Recognition): Goal-Recognition is the task to recognize the goal...
% Definition of Types.
% Location includes all the blocks and the table.
location(X) :- block(X). 
location(table).
% Definition of Fluents.
fluent(inertial, on(B,L)) :- block(B), location(L). 
fluent(inertial, clear(B)) :- block(B). 
% Definition of Actions.
action(move(B,L1,L2)) :- block(B), location(L1), location(L2), 
    B!=L1, B!=L2, L1!=L2.
% Action effects.
holds(on(B,L2),I+1) :- occurs(move(B,L1,L2),I), I < num_step.
-holds(on(B,L1),I+1) :- occurs(move(B,L1,L2),I), I < num_step.
holds(clear(L1),I+1) :- occurs(move(B,L1,L2),I), I < num_step.
-holds(clear(L2),I+1) :- occurs(move(B,L1,L2),I), I < num_step.
% Action precondition.
-occurs(move(B,L1,L2),I) :- location(L1), location(L2), holds(on(B1,B),I). 
-occurs(move(B,L1,L2),I) :- location(L1), location(L2), -holds(clear(B),I). 
-occurs(move(B1,L1,B2),I) :- block(B1), location(L1), block(B2), holds(on(B,B2),I).
-occurs(move(B1,L1,B2),I) :- block(B1), location(L1), block(B2), -holds(clear(B2),I).
-occurs(move(B,L1,L2),I) :- location(L2), -holds(on(B,L1),I). 
% General inertia axiom.
holds(F,I+1) :- fluent(inertial ,F), holds(F,I), not -holds(F,I+1), I<num_step. 
-holds(F,I+1) :- fluent(inertial ,F), -holds(F,I), not holds(F,I+1), I<num_step. 
% Close world assumption for Actions.
-occurs(A,I) :- action(A), step(I), not occurs(A,I).
% Close world assumption for init state.
-holds(on(B,L),0) :- block(B), location(L), not holds(on(B,L),0).
-holds(clear(B),0) :- block(B), not holds(clear(B),0).
% Do not allow concurrent actions.
:- action(A1), action(A2), occurs(A1,I), occurs(A2,I), A1 != A2.
% Rules for planning.
success :- goal(I), I<=num_step. 
:- not success. 
occurs(A,I) | -occurs(A,I) :- action(A), step(I), not goal(I), I<num_step. 
% Rule for optimal plan. Occurred (Action, K) should be minimized.
#minimize{1, Action, K: occurs(Action , K)}.

......(few-shot examples for other Blocks Would domain puzzles)

% Introduction: Domain: We use a variant of the blocks world (BW) as an example where...
% Task (Legality): This task directly targets the preconditions of actions...
% Definition of Types.
% Location includes all the blocks and the table.
ASP_RULES
% Definition of Fluents.
ASP_RULES
% Definition of Actions.
ASP_RULES
% Action effects.
ASP_RULES
% Action precondition.
ASP_RULES
% General inertia axiom.
ASP_RULES
% Close world assumption for Actions.
ASP_RULES
% Close world assumption for init state
ASP_RULES
% Do not allow concurrent actions: 
ASP_RULES

Complete the ASP_RULES above. Do not add new rules.
\end{verbatim}
\end{tcolorbox}
\caption{Prompts for learning rule specifications in the Blocks World - Legality puzzles. We first present few-shot examples from other puzzles from the Blocks World domain. Then we prompt the LLM to translate rule specifications from natural language to ASP programs.}
\label{fig:fs-rule-bw}
\end{figure*}

\begin{figure*}
\centering    
\begin{tcolorbox}[colback=gray!5, colframe=black,fontupper=\small]
\begin{verbatim}
State Input:
The lime block is on top of the aquamarine block. The olive block is clear. The lime block 
is clear. The teal block is on the table. The olive block is on the table. The navy block 
is on top of the teal block. The navy block is clear. The aquamarine block is on the table.
Query Input:
Jane moves the navy block from the teal block to the lime block. Jane moves the navy block 
from the lime block onto the table. Jane moves the lime block from the aquamarine block to 
the navy block.
% State Encoding:
holds(on(lime, aquamarine), 0).
holds(clear(olive), 0).
holds(clear(lime), 0).
holds(on(teal, table), 0).
holds(on(olive, table), 0).
holds(on(navy, teal), 0).
holds(clear(navy), 0).
holds(on(aquamarine, table), 0).
% All Blocks:
block(lime). block(olive). block(teal). block(navy). block(aquamarine).
% Query Encoding:
occurs(move(navy, teal, lime), 0).
occurs(move(navy, lime, table), 1).
occurs(move(lime, aquamarine, navy), 2).
% Steps:
#const num_step=3.

State Input:
{state}
Query Input:
{query}

......(other few-shot examples)

Complete the text, output % State Encoding, % State Encoding, % All Blocks, 
% Query Encoding, % Steps.
Only show the result, do not explain.
\end{verbatim}
\end{tcolorbox}
\caption{Prompts for translating state representations to ASP programs in the Blocks World - Legality puzzle.}
\label{fig:fs-state-bw}
\end{figure*}

\begin{figure*}
\centering    
\begin{tcolorbox}[colback=gray!5, colframe=black,fontupper=\small]
\begin{verbatim}
You are a reasoning assistant in a blocks world.
Rules:
1. All blocks are of equal size.
2. A block can only be moved if it is clear (nothing on top of it).
3. A block can only be placed on another block if the target block is clear.
4. Blocks may also be placed on the table, which always has space.
5. A block can only be on one other block or the table.

Decide whether a given action is legal in the initial state.

Input format:
state: A set of facts describing the world.
query: An action to evaluate.

Output:
Format your answer starting with % followed by a line of either True or False.

Example:
Input
state:
The red block is on the table. The tan block is on the table. The navy block is on top of 
the turquoise block. The tan block is clear. The red block is clear. The green block is 
clear. The green block is on the table. The navy block is clear. The turquoise block is on 
the table.
query:
Jane moves the navy block from the turquoise block to the green block.

Output:
True

The new input is as follows:
Input:
state: 
{state}

query: 
{query}
\end{verbatim}
\end{tcolorbox}
\caption{A standard prompt for solving the Blocks World - Legality puzzle.}
\label{fig:standard-bw}
\end{figure*}

\end{document}